\pdfoutput=1

\documentclass[11pt]{article}

\usepackage[]{emnlp2021}

\usepackage{times}
\usepackage{latexsym}

\usepackage{mathtools}
\usepackage{booktabs,arydshln}
\usepackage{multirow}
\usepackage{adjustbox}
\usepackage{arydshln}
\usepackage{tabularx}
\usepackage{array}
\usepackage{graphicx,subcaption}
\usepackage{capt-of}
\usepackage{mathabx}

\usepackage[ruled,vlined,linesnumbered]{algorithm2e}
\usepackage{etoolbox}

\usepackage{xspace}

\usepackage[T1]{fontenc}

\usepackage[utf8]{inputenc}

\usepackage{microtype}

\title{Cross-Lingual Training with Dense Retrieval for Document Retrieval}

\author{Peng Shi\textsuperscript{\rm 1},
  Rui Zhang\textsuperscript{\rm 2},
  He Bai\textsuperscript{\rm 1}, and Jimmy Lin\textsuperscript{\rm 1} \\
\textsuperscript{\rm 1} David R. Cheriton School of Computer Science, University of Waterloo\\
\textsuperscript{\rm 2} Department of Computer Science and Engineering, Penn State University \\
{\tt \{peng.shi,he.bai,jimmylin\}@uwaterloo.ca,
 \tt rmz5227@psu.edu} 
}

\begin{document}
\maketitle
\begin{abstract}

Dense retrieval has shown great success in passage ranking in English.
However, its effectiveness in document retrieval for non-English languages remains unexplored due to the limitation in training resources.
In this work, we explore different transfer techniques for document ranking from English annotations to multiple non-English languages.
Our experiments on the test collections in six languages~(Chinese, Arabic, French, Hindi, Bengali, Spanish) from diverse language families reveal that
zero-shot model-based transfer using mBERT improves the search quality in non-English mono-lingual retrieval. 
Also, we find that weakly-supervised target language transfer yields competitive performances against the generation-based target language transfer that requires external translators and query generators.

\end{abstract}

\section{Introduction}
Dense retrieval uses dense vector representations for semantic encoding and matching.
It has shown its effectiveness in open-domain question answering and passage ranking~\cite{nogueira2019passage,yang2019simple,karpukhin2020dense}.
However, most of the existing work focus on high-resource languages such as English, where large-scale annotations are readily accessible.
Widely available large-scale datasets such as Natural Questions~(NQ)~\cite{kwiatkowski2019natural} and MS MARCO~\cite{nguyen2016ms} are used for training dense retrieval encoder to achieve state-of-the-art performances in English.
Such data is especially hard to obtain for low-resource languages, considering that large amounts of annotations are required for training dense retrieval encoders.

More recently, \citet{shi2020cross} proposed several cross-lingual transfer techniques for bridging the language gaps between high-resource languages and low-resource languages for document reranking.
However, unlike dense retrieval, document reranking affects the efficiency of the document retrieval because reranking is the second stage inference based on the results of first-stage term-matching based retrieval.

In this work, we focus on improving the traditional term-matching based retrieval without losing the efficiency of the retriever.
Our model combines term-matching scores and dense retrieval similarity in a single step, overcoming the efficiency issue of two-stage document reranking and exploiting the representation ability of neural models at the same time.
More specifically, we explore techniques for leveraging the relevance judgments in a source language, usually a high-resource language such as English, to train dense retrievers for mono-lingual document retrieval in multiple target~(non-English) languages.
Note that this setting is different from \textit{cross-lingual information retrieval}~(CLIR), where queries and documents are in different languages~\cite{shi2019cross,yu2020study,jiang2020cross,litschko2021evaluating,yu2021cross,chen2021cross}.
Following the idea of cross-lingual training~\cite{shi2020cross}, we examine low-resource techniques including zero-shot model-based transfer and weakly-supervised target language transfer; 
we also explored the technique by leveraging public translators, such as Google Translate, to improve the language transfer of the dense retriever.

We train our retrieval models using Natural Questions and MS MARCO in English, and we evaluate our transfer approaches on diverse text collections for document retrieval in six target languages including NTCIR 8 in Chinese~\cite{mitamura2008overview}, TREC 2002 in Arabic~\cite{oard2002trec}, CLEF 2006 in French~\cite{di2006clef}, FIRE 2012 in Hindi, FIRE 2012 in Bengali~\cite{yadavism}, TREC 3 in Spanish~\cite{harman1995overview}.
Our experimental results show that while the dense retriever itself cannot obtain performances as competitive as traditional exact term matching methods such as BM25 or BM25+RM3 query expansion, combining them can obtain effectiveness improvements without losing the efficiency.
Especially, the weakly-supervised target language transfer yields competitive performances compared with the generation-based target language transfer that requires external translators.

\section{Related Work}
Dense retrieval showed its superiority over the traditional term matching based methods such as BM25 or BM25+RM3 query expansion on passage retrieval task~\cite{dai2019deeper,dai2019context,karpukhin2020dense,zhang2019improving,chang2020pre,macavaney2020teaching}.
A bi-encoder architecture is used for the dense retrievers, where the queries and documents are mapped into hidden vectors independently without any interaction between them.
Compared with term-based sparse retrieval using TF-IDF or BM25, it can capture synonyms or paraphrases by incorporating contexts and provide additional flexibility to learn task-specific representations~\cite{lin2020pretrained}.

Document reranking is another topic in document retrieval.
With pretrained language models, the reranking effectiveness has been improved significantly~\cite{nogueira2019passage, yilmaz2019applying}.
A cross-encoder architecture is often used for document reranking, where the tokens of query and the document can have full interactions in the encoder and the matching signals can be easily captured by the models.

The multilingual BERT~\cite{devlin2018bert} has shown its language transfer abilities over different tasks~\cite{wu2019beto}.
\citet{shi2020cross} were the one of the first to build IR re-rankers based on the mBERT for non-English corpus by leveraging the relevance judgments in English.

\section{Preliminaries}
We explore different strategies for the cross-lingual transfer of dense passage retriever~(DPR) to improve traditional retrieval methods such as BM25 or BM25+RM3.
DPR uses a dense encoder $E_{P}(\cdot)$ that encodes passages to a $d$-dimension vectors and builds an index for document collections that are used for retrieval.
For the query, a different encoder $E_{Q}(\cdot)$ is used for mapping the query into $d$-dimension vector.
Based on the query vector, the closest top $k$ passages are retrieved from the pre-built index based on the pre-defined distance function.
Following~\citet{karpukhin2020dense}, we define the distance function as $sim(q, p) = E_{Q}(q)^\top E_{P}(p)$.
We use pretrained mBERT as the start point and the hidden state of \texttt{[CLS]} token is regarded as the representation of the encoded text.
During inference, we apply both bag-of-words exact term matching such as BM25 or BM25+RM3 and dense retrieval.
The relevance score of each document combines the term-matching scores with dense retrieval similarity 
$S_{doc} = \alpha \cdot S_{term} + (1-\alpha) \cdot S_{dense}$ 
where $\alpha$ is tuned via cross-validation.
All candidate documents are sorted by the above score $S_{doc}$ to produce the final output.

\section{Cross-Lingual Relevance Transfer}
To perform cross-lingual transfer of DPR from a high-resource source language to low-resource target languages, we investigate two groups of strategies.
The first strategy, model-based transfer, directly applies the retrieving model trained on the source language to other target languages in a zero-shot manner.
The second strategy explores two data augmentation techniques to build the training data on target languages for finetuning.

\subsection{Model-based Transfer}
By exploiting the zero-shot cross-lingual transfer ability of pre-trained transformers such as mBERT~\cite{devlin2018bert} and XLM-Roberta~\cite{conneau2019unsupervised}, 
we train the dense retriever encoder in the source language and apply inference directly on target languages.
These pre-trained transformers only require raw text in different languages, e.g. Wikipedia, and are trained in a self-supervised manner, so we characterize this approach as ``low resource''.

\subsection{Target Language Transfer}
To bridge the language gap between the training and the inference, a direct solution is target language data augmentation.
In this work, we explore two techniques for creating a target language transfer set, including generation-based query synthesis and weakly supervised query synthesis. 

\paragraph{Generation-based Query Synthesis.}
The goal of the generation-based query synthesis is to leverage powerful generation models to predict reasonable queries given documents in the target language.
We choose the multilingual version of BART~(mBART)~\cite{liu2020multilingual}, a pretrained sequence-to-sequence transformers, as our query generation model.
The input of the model is the passage and its learning target is the corresponding query.
We use the translate-train technique to obtain the generation models.
More specifically, we leverage Google Translate to translate the query-document pairs in English to target languages.
In the inference stage, we use the passages in the target language collections as the input and generate corresponding queries in the same language.
In our preliminary experiments, we also tried zero-shot transfer that model is trained on English query-document pairs and directly applied to target languages for query inference.
However, the generated queries are of low quality, and this observation is also confirmed by~\citet{chi2020cross}.

\smallskip \noindent \textbf{Weakly-supervised Query Synthesis.}
Wikipedia has documents in varies languages, and it is a good transfer set in the cross-lingual training.
We can automatically build the target language transfer set without manual annotation effort, by treating the titles of Wikipedia articles as queries and the corresponding documents as positive candidates.
We also retrieve top 1000 documents with BM25 for each query, and the documents except the positive candidate are labeled as negative candidates.
Queries whose positive document is not in the retrieved set are removed.
In this way, we can obtain query-document pairs in target languages.

\paragraph{Two-stage Training.}
We apply two-stage training to train the dense retriever encoders.
The dense retriever encoders are firstly trained on source language annotated data which are available in a large scale;
then the models are finetuned on the synthesized query-document pairs in the target language.

\begin{table}[t]
    \centering
    \resizebox{\columnwidth}{!}{%
		\begin{tabular}{llrr}
			\toprule
			Doc Language & Source & \# Topics & \# Docs \\
			\midrule
			Chinese& NTCIR 8 & 73 & 308,832 \\
			Arabic & TREC 2002 & 50 &383,872 \\
			French & CLEF 2006 & 49 & 171,109 \\
			Hindi & FIRE 2012 & 50 & 331,599 \\
			Bengali & FIRE 2012 & 50 & 500,122 \\
			Spanish & TREC 3 & 25 &  57,868 \\
		    \bottomrule
	\end{tabular}
	}
	\caption{Dataset statistics for test collections.}
	\label{table:dataset}
	
\end{table}

\begin{table*}[t!]
    \small
    \centering
    \resizebox{\textwidth}{!}{%
  \begin{tabular}{l lll lll lll}
  \toprule
    &  {\bf AP} & {\bf P@20} & {\bf nDCG} & {\bf AP}  & {\bf P@20} & {\bf nDCG} & {\bf AP} & {\bf P@20} & {\bf nDCG} \\
    \cmidrule(lr){2-4}  \cmidrule(lr){5-7}  \cmidrule(lr){8-10}
    {\bf Model} & \multicolumn{3}{c}{\textbf{NTCIR8-zh}} & \multicolumn{3}{c}{\textbf{TREC2002-ar}} & \multicolumn{3}{c}{\textbf{CLEF2006-fr}} \\
    \toprule
    $ {\bf (0)}~\textrm{BM25} $ & $0.4014^{}$ & $0.3849^{}$ & $0.4757^{}$ & $0.2932^{}$  & $0.3610^{}$ & $0.4056^{}$ & $0.3111^{}$ & $0.3184^{}$ & $0.4458^{}$ \\
    $ {\bf (1)}~\textrm{BM25+RM3} $ & $0.3384^{}$ & $0.3616^{}$ & $0.4490^{}$  & $0.2783^{}$ & $0.3490^{}$ & $0.3969^{}$ & $0.3421^{}$ & $0.3408^{}$ & $0.4658^{}$ \\
    \cmidrule(lr){1-1} \cmidrule(lr){2-4} \cmidrule(lr){5-7}  \cmidrule(lr){8-10}
    $ {\bf (2)}~\textrm{NQ zero-shot}$ & $0.4221^{\blacktriangleup}$ & $0.4164^{\blacktriangleup}$ & $0.5235^{\blacktriangleup}$ & $0.2943^{}$ & $0.3560^{}$ & $0.4012^{}$ & $0.3470^{}$ & $0.3469^{}$ & $0.4726^{}$ \\
    $ {\bf (3)}~\textrm{MS zero-shot}$ & $0.4167^{\blacktriangleup}$ & $0.4164^{\blacktriangleup}$ & $0.5095^{\blacktriangleup}$ & $0.3024^{}$ & $0.3810^{\blacktriangleup}$ & $0.4285^{}$ & 
    $0.3332^{}$ & $0.3418^{}$ & $0.4573^{}$ \\
    \cmidrule(lr){1-1} \cmidrule(lr){2-4} \cmidrule(lr){5-7}  \cmidrule(lr){8-10}
    $ {\bf (4)}~\textrm{MS} \rightarrow \textrm{QGen}$ & $0.4258^{\blacktriangleup}$ & $0.4336^{\blacktriangleup}$ & $0.5308^{\blacktriangleup}$ & $0.2988^{}$ & $0.3800^{}$ & $0.4276^{}$ & $0.3331^{}$ & $0.3429^{}$ & $0.4564^{}$ \\
    $ {\bf (5)}~\textrm{MS} \rightarrow \textrm{Wiki}$ & $0.4135^{}$ & $0.4123^{\blacktriangleup}$ & $0.5055^{\blacktriangleup}$ & $0.3060^{\blacktriangleup}$ & $0.3750^{}$ & $0.4293^{}$ & $0.3456^{}$ & $0.3480^{}$ & $0.4743^{}$ \\
    \midrule
    & \multicolumn{3}{c}{\textbf{FIRE2012-hi}} & \multicolumn{3}{c}{\textbf{FIRE2012-bn}} & \multicolumn{3}{c}{\textbf{TREC3-es}} \\
    \toprule
    $ {\bf (0)}~\textrm{BM25} $ & $0.3867^{}$ & $0.4470^{}$ & $0.5310^{}$ & $0.2881^{}$  & $0.3740^{}$ & $0.4261^{}$ & $0.4197^{}$ & $0.6660^{}$ & $0.6851^{}$ \\
    $ {\bf (1)}~\textrm{+RM3} $ & $0.3660^{}$ & $0.4430^{}$ & $0.5277^{}$ & $0.2833^{}$ & $0.3830^{}$ & $0.4351^{}$ & $0.4912^{}$ & $0.7040^{}$ & $0.7079^{}$ \\
    \cmidrule(lr){1-1} \cmidrule(lr){2-4} \cmidrule(lr){5-7}  \cmidrule(lr){8-10}
    ${\bf (2)}~\textrm{NQ zero-shot}$ & $0.3939^{}$ & $0.4560^{}$ & $0.5408^{}$ & $0.2898^{}$ & $0.3980^{}$ & $0.4495^{\blacktriangleup}$ & $0.4910^{}$ & $0.6980^{}$ & $0.7007^{}$ \\
    $ {\bf (3)}~\textrm{MS zero-shot}$ & $0.3944^{}$ & $0.4580^{}$ & $0.5461^{}$ & $0.2896^{\blacktriangleup}$ & $0.3900^{}$ & $0.4449^{}$ & $0.4950^{}$ & $0.7080^{}$ & $0.7171^{}$ \\
    \cmidrule(lr){1-1} \cmidrule(lr){2-4} \cmidrule(lr){5-7}  \cmidrule(lr){8-10}
    $ {\bf (4)}~\textrm{MS} \rightarrow \textrm{QGen}$ & $0.3941^{}$ & $0.4660^{}$ & $0.5527^{}$ & $0.2887^{}$ & $0.3980^{}$ & $0.4486^{}$ & $0.4958^{\blacktriangleup}$ & $0.7180^{}$ & $0.7239^{}$ \\
    $ {\bf (5)}~\textrm{MS} \rightarrow \textrm{Wiki}$ & $0.3950^{}$ & $0.4630^{}$ & $0.5497^{}$ & $0.2898^{\blacktriangleup}$ & $0.4050^{}$ & $0.4549^{}$ & $0.4972^{\blacktriangleup}$ & $0.7180^{}$ & $0.7329^{}$ \\
    \bottomrule
    \end{tabular}
	}
	\caption{Experimental results on the baselines and our cross-lingual transfer methods. Model (0) and (1) are term matching baselines. Model (2) NQ zero-shot: zero-shot transfer of models trained on Natural Questions. Model (3) MS zero-shot: zero-shot transfer of models trained on MS MARCO. Model (4) $\textrm{MS} \rightarrow \textrm{QGen}$: trained on MS MARCO and then finetuned on query generation data requiring external translators. Model (5) $\textrm{MS} \rightarrow \textrm{Wiki}$: trained on MS MARCO and then finetuned on query synthesis data from Wikipedia. For nDCG, we report nDCG@20. Significant gains against the baselines are denoted with $\blacktriangleup$.}
	\label{table:multilingual}
\end{table*}

\section{Experimental Setup}

\paragraph{Evaluation.}
We conduct experiments on six test collections in diverse languages~(Chinese, Arabic, French, Hindi, Bengali, Spanish).
Data statistics are shown in Table~\ref{table:dataset}.
For the evaluation metrics, we adopt the average precision~(AP) up to rank 1000, 
precision at rank 20~(P@20) and nDCG at rank 20~(nDCG), computed with the \texttt{trec\_eval} toolkit.
The query was used to retrieve the top 1000 hits from the corpus using BM25 or BM25+RM3 query expansion; the default Anserini~\cite{yang2017anserini} settings were used.
For the dense retrieval, the top 100 hits are retrieved from the index and the similarity scores are combined with the term-matching scores.
For the documents that are not in the retrieved set, either from term-matching methods or dense retrieval, 0 score is assigned.
Fisher’s two-sided, paired randomization test~\cite{smucker2007comparison} at $p < 0.05$ was applied to test for statistical significance.

\paragraph{Training.}
For the model-based transfer, we explore two training datasets in the source language, English in our case, including the Natural Question and MS MARCO.
Note that Natural Question is an open-domain question answering dataset, 
where the queries are usually long questions instead of a bag of keywords in the document ranking datasets. 
This introduces a new gap in query style besides language in the transfer process. 
For training the query generator in target languages, we obtain the training data by sampling 2000 query-passage pairs from MS MARCO and translate them into target languages same as target benchmarks.
For two-stage training, we first train the dense encoders on the MS MARCO dataset and then further tune them on the transfer set.

\paragraph{Inference.}
For dense retrieval, documents are often too long to fit into BERT models for encoding.
A common approach to address this issue is to split the long document into segments within fixed length~(e.g. 512 tokens), and build an index based on the segments.
In this work, we segmented the documents using a sliding window of 5 sentences and a stride of 1 sentence. 
For each query, the score of document $S_{dense}$ is obtained by averaging the top-3 scores of the retrieved segments.
We applied five-fold cross-validation on all datasets, choosing parameter $\alpha$ that yielded the highest AP.

\section{Results and Discussions}

Our results are shown in Table~\ref{table:multilingual}.
Models (0) and Models (1) show the effectiveness of BM25 and BM25 with RM3 query expansion.
For each language, we select the higher P@20 of the two models as the term-based matching baselines.
That is, for the French, Bengali and Spanish collections, we use the BM25+RM3 as the term-based matching baseline and for others, we use the BM25.

\paragraph{Finding \#1: Zero-shot model-based transfer improves term-based matching.}
The results of zero-shot model-based transfer are shown in Models~(2) and Models~(3).
Comparing with the corresponding baselines, we observe that the model-based transfer, either NQ zero-shot or MS zero-shot, can improve the retrieval effectiveness on P@20 for all collections, except the NQ zero-shot on TREC3-es dataset.
We do not observe a clear winner between NQ and MS, though.
For example, Models~(2) perform better on Chinese and French collections; while Models~(3) yield better retrieval effectiveness on Arabic and Spanish collections.
Since mBERT is widely available, mono-lingual retrieval improvements can be obtained “for free” with annotated data in English. 
These results indicate that mBERT-based DPR effectively transfers relevance matching across languages. 

\paragraph{Finding \#2: Target language transfer benefits certain collections, and Wiki query synthesis is better than query generation.}
Target language transfer results are shown in Models~(4) and Models~(5).
$\textrm{MS} \rightarrow \textrm{QGen}$ and $\textrm{MS} \rightarrow \textrm{Wiki}$ denote two-stage training strategy with different transfer sets, 
where $\textrm{QGen}$ denotes generation-based query synthesis and $\textrm{Wiki}$ denotes weakly supervised query synthesis from Wikipedia.
By comparing the Models~(4) with Models~(3), we observe the second stage training with generation-based query-document pairs  
can improve the effectiveness of P@20 over the zero-shot model-based transfer on Chinese, French, Hindi, Bengali and Spanish collections.
However, there is no difference over AP for all collections.
By comparing the Models~(5) with Models~(3), we find that the second stage training with weakly-supervised training data can improve the P@20 over the zero-shot baselines on French, Hindi, Bengali and Spanish collections. 

Furthermore, by comparing these two transfer sets, we observe that, except for the Chinese collection, the $\textrm{Wiki}$ obtains better retrieval effectiveness than $\textrm{QGen}$, which requires external translators and training query generators on different languages.

\section{Conclusion}
We investigate the effectiveness of three transfer techniques for document ranking from English training data to low-resource target languages.
Our experiments in six languages demonstrate that zero-shot transfer of mBERT-based dense retrieval models improves traditional term-based matching method, and finetuning on augmented data in target languages can further benefit certain collections.

\bibliography{anthology,custom}
\bibliographystyle{acl_natbib}

\end{document}